\documentclass[times, 10pt,twocolumn]{article} 
\usepackage{latex8}
\usepackage{times}
\usepackage[linesnumbered,ruled,vlined]{algorithm2e}
\usepackage{graphicx}
\usepackage{caption}
\usepackage{subcaption}
\usepackage{amsmath}
\usepackage{hyperref}
\begin{document}

\title{2D Grid Map Generation for Deep-Learning-based Navigation Approaches}
%\title{Two-dimensional map generator algorithm for indoor navigation}

\author{Gabriel O. Flores-Aquino\\
gfloresa0500@alumno.ipn.mx\\
Instituto Polit\'ecnico Nacional (IPN)\\ 
SEPI-UPIITA, M\'exico City, M\'exico\\ 
\and
Jheison Duvier Díaz Ortega\\
jdiazo1800@alumno.ipn.mx\\
Instituto Polit\'ecnico Nacional (IPN)\\ 
SEPI-UPIITA, M\'exico City, M\'exico\\ 
\and
Ricardo Yahir Almazan Arvizu\\
ralmazana1700@alumno.ipn.mx\\
Instituto Polit\'ecnico Nacional (IPN)\\ 
SEPI-UPIITA, M\'exico City, M\'exico\\ 
\and
Ra\'ul L\'opez Mu\~{n}oz\\
rlopezm1209@alumno.ipn.mx\\
Instituto Polit\'ecnico Nacional (IPN)\\
CIDETEC, M\'exico City, M\'exico\\ 
\and
O. Octavio Gutierrez-Frias\\
ogutierrezf@ipn.mx\\
Instituto Polit\'ecnico Nacional (IPN)\\ 
SEPI-UPIITA, M\'exico City, M\'exico \\ 
\and
J. Irving Vasquez-Gomez\\
jvasquezg@ipn.mx\\
Instituto Polit\'ecnico Nacional (IPN)\\
CIDETEC, M\'exico City, M\'exico\\ 
}

%\author{Gabriel O. Flores-Aquino\\
%SEPI-UPIITA, Instituto Polit\'ecnico Nacional\\ 
%M\'exico City, M\'exico\\ gfloresa0500@alumno.ipn.mx\\
%\and
%Jheison Duvier Díaz Ortega\\
%SEPI-UPIITA, Instituto Polit\'ecnico Nacional\\ 
%M\'exico City, M\'exico\\ jdiazo1800@alumno.ipn.mx\\
%\and
%Ricardo\\
%SEPI-UPIITA, Instituto Polit\'ecnico Nacional\\ 
%M\'exico City, M\'exico\\ jdiazo1800@alumno.ipn.mx\\
%\and
%Raúl López Muñoz\\
%CIDETEC, Instituto Polit\'ecnico Nacional\\
%M\'exico City, M\'exico\\ rlopezm1209@alumno.ipn.mx\\
%\and
%J. Irving Vasquez-Gomez\\
%CIDETEC, Instituto Polit\'ecnico Nacional\\
%M\'exico City, M\'exico\\ jvasquezg@ipn.mx\\
%\and
%Octavio Gutierrez-Frias\\
%SEPI-UPIITA, Instituto Polit\'ecnico Nacional\\
%M\'exico City, M\'exico\\ ogutierrezf@ipn.mx\\
%}

\maketitle
\thispagestyle{empty}

\begin{abstract}
In the last decade, autonomous navigation for robotics has been leveraged by deep learning and other approaches based on machine learning. These approaches have demonstrated significant advantages in robotics performance. But they have the disadvantage that they require a lot of data to infer knowledge. In this paper, we present an algorithm for building 2D maps with attributes that make them useful for training and testing machine-learning-based approaches. The maps are based on dungeons environments where several random rooms are built and then those rooms are connected. In addition, we provide a dataset with 10,000 maps produced by the proposed algorithm and a description with extensive information for algorithm evaluation. Such information includes validation of path existence, the best path, distances, among other attributes. We believe that these maps and their related information can be useful for robotics enthusiasts and researchers who want to test deep learning approaches. The dataset  is available at \url{https://github.com/gbriel21/map2D_dataSet.git}.

%\href{https://github.com/gbriel21/map2D_dataSet.git}{github}.
\end{abstract}

%------------------------------------------------------------------------- 
\Section{Introduction}

\begin{figure}[tb]
    \centering
    \includegraphics[width=\linewidth]{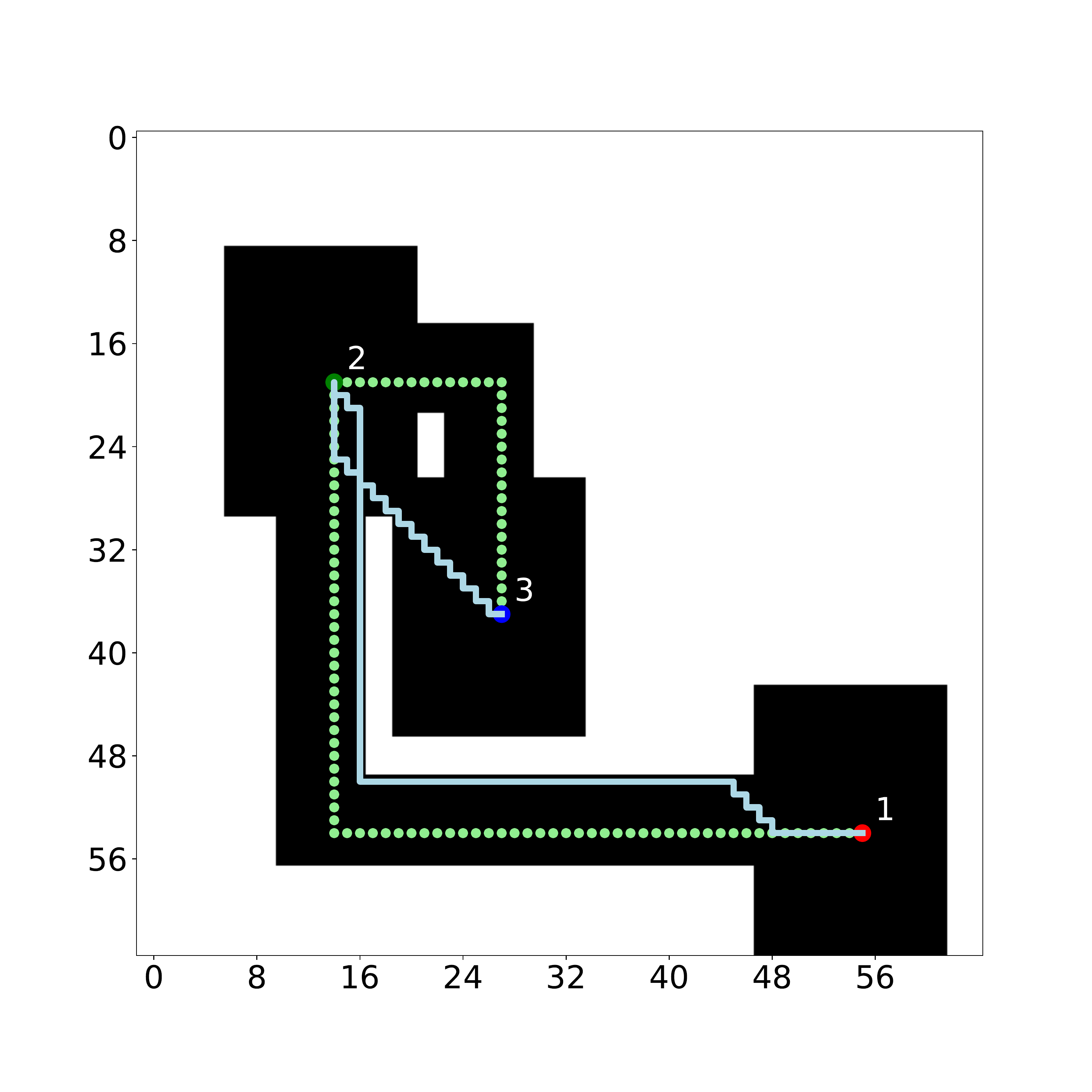}
    \caption{Example of a map built by the proposed method. The illustration shows three random rooms that are connected by corridors as well as the paths than connect those rooms. Free cells are drawn in black. Occupied cells are drawn in white. In green dotted lines, breath-first search path is drawn. In solid light blue line A* computed path is drawn.}
    \label{fig:map497}
\end{figure}
%\noindent

Our experience in the human world dictates to us that the best way to interact with an environment is by using local and global references. This idea is used in Map-based navigation where the robot knows a priory a representation of the work-space. This representation, denominated map, is usually encoded as an occupancy grid where each grid represented an unoccupied or occupied space \cite{Corke}. The occupancy grid maps are widely used for probabilistic localization and mapping \cite{Thrun} as well for planning algorithms as Rapidly Random Trees (RRT) \cite{RRT}, RRT* \cite{RRTstar} or Probabilistic Road Maps (PRM) \cite{PRM} to name a few classic examples. The occupancy maps are also necessary for new approaches. For example, in \cite{TowardAutonomous}, where the input of a neural network is a dungeon map. In \cite{motionPlanningNetworks}, the authors describe a neuronal planner for solving motion problems. This approach required a large amount of data including, maps or environments useful for testing and training. Another recent use for maps is to infer a latent representation of the work-space, a good example is exposed in \cite{planningLatentSpaces}. Other recent works explore the uses of reinforcement learning or deep reinforcement learning, such as \cite{PRMRL}, where combines sampling-based planning with reinforcement learning. In this work, planning is tested in large spaces represented as an occupancy grid, but the training process is built with small environments. Other works use other learning approaches such as supervised learning techniques. An example of this approach is presented in \cite{lego}, where a neural network improves the sampling in difficult passages using as input a map.

Recent and unexpected changes in the world such as those caused by the Covid 19 pandemic have accelerated the trend of using service robots for tasks in home and workspaces as hospitals or offices to guarantee safety measures. For these reasons, the proposed maps are intended to represent indoor spaces, similar to dungeons maps used in a large number of computer games \cite{dungeons}, composed of square areas like rooms in a house or office spaces and connected by hallways.

In this paper, we propose an algorithm to generate random bi-dimensional maps. The algorithm can be customized, by modifying simple parameters, for example, the size of the output image or the number of rooms. Besides, to increase the diversity and complexity of the maps, the proposed algorithm has several random variables that allow created maps with different sizes for rooms and hallways and maps fully connected as maps with inaccessible regions. Together with this algorithm, we also provide an associated dataset with 10,000 images complemented with a CSV file with information about the parameters that characterize the maps and paths founded with the algorithms, Breadth-first search \cite{IntroductionToAlgorithms} and A* search \cite{Astar} (see Figure \ref{fig:map497}) and a connectivity matrix that provides information about connection between rooms. We believe that the dataset and associated information can be useful in machine learning approaches, especially for indoor navigation. 

The document is organized as follows. In section \ref{sec_algorithm}, we present the algorithm to generate maps. In section \ref{sec_maps}, we describe the maps obtained, and in section \ref{sec_makeup}, we explain the characterization for maps. Finally, in section \ref{sec_conclusion}, we provide our conclusions and future work.

%------------------------------------------------------------------------- 
\Section{Map Generation Algorithm}
\label{sec_algorithm}
In this section, we present the pseudo-code for building $k$ bi-dimensional maps (see Algorithm \ref{alg:one}). Where each map is conformed as a set of square rooms connected with square passages. The input is the number of maps required and the output is a set of maps with the characteristics described in the parameters. These parameters are \textbf{map size} in pixels; \textbf{number of rooms} with a minimum of two rooms, minimum and maximum limits for the room (\textbf{room width limit, room length limit}), and \textbf{tunnel width limit} for the amplitude of the passages.

\begin{algorithm}
\caption{Map 2D}\label{alg:one}
\SetKwInOut{Input}{Input}
\SetKwInOut{Output}{Output}
\SetKwInOut{Parameters}{Parameters}

\Input{Number of maps $:k$}
\Output{Set of maps: $maps$}
\Parameters{\\
Map size $:(x,y)$,\\ 
Number of rooms $:n$,\\ 
Room width limit$ : [w_{min},w_{max}]$,\\%columns 
Room length limit$ : [l_{min},l_{max}]$,\\%rows
Tunnel width limit$: [t_{min},t_{max}]$}
\BlankLine
%Create an array of size
\While{$m\leq k$}{
    $Mtx(x,y) \gets 1$;\\%all occuped 
    $R(2,n) \gets 0$;\\
    $T \gets rand(t_{min},t_{max})$;\\
    \For{$p=1$ \KwTo $n$}{
        %create center point
        $c,f \gets rand(0,x),rand(0,y)$;\\
        $R[0,p] \gets c$;\\
        $R[1,p] \gets f$;\\
        $r_{w} \gets rand([w_{min},w_{max}])$;\\
        $r_{l} \gets rand([l_{min},l_{max}])$;\\
        $Mtx[c-\frac{r_{w}}{2}:c+\frac{r_{w}}{2},f-\frac{r_{l}}{2}:f+\frac{r_{l}}{2}] \gets 0$}
    %create tunnels
    \For{q=1 \KwTo n-1}{
        $connect \gets rand(0,10)$;\\
        \eIf{$connect != 1$}{
            $A \gets R[0,q],R[1,q]$;\\
            $B \gets R[0,q+1],R[1,q+1]$;\\
           $direction \gets rand(0,1)$;\\
           \eIf{direction==1}{
           \tcc{connect A horizontal and B vertical}
           $Mtx[A[0]:B[0],A[1]-\frac{T}{2}:A[1]+\frac{T}{2}]\gets 0$;\\
           $Mtx[B[0]-\frac{T}{2}:B[0]+\frac{T}{2},A[1]:B[1]]\gets 0$}
           
           {\tcc{connect A vertical and B horizontal}
           $Mtx[A[0]-\frac{T}{2}:A[0]+\frac{T}{2},A[1]:B[1]]\gets 0$;\\
           $Mtx[B[1]-\frac{T}{2}:B[1]+\frac{T}{2},A[0]:B[0]]\gets 0$}
            }
        {do not connect;}
        }
$m++$;\\
$maps  \gets$ store ($Mtx$);\\
return $maps$
}
\end{algorithm}

\par
In line 2, we define a matrix of size (x, y); we consider 1 as an occupied space and 0 as free space, in line 3 we define and array to store the seed points of each room and in line 4 we choose randomly a value for the width of the tunnel. In the first part, we free up the space of the rooms (\textbf{lines 5} to \textbf{11}). In \textbf{line 5} for $p=\{1,2,... ...,n\}$ where $n$ is the number of rooms, we randomly created $n$ seeds or origin points (\textbf{line 6}) and save them in the array $R$. Subsequently, in \textbf{lines 9} and \textbf{10} we get randomly, the width and length for the n-th element. For the second part, we need to connect each pair of rooms. To add complexity in \textbf{line 13}, we add the option of not connecting a couple of rooms in the ten percent of cases. In \textbf{lines 15} and \textbf{16}, we select progressively a pair of rooms denoted as A and B to connect in one of two ways when A is a horizontal passage and, consequently, B is a vertical passage, \textbf{lines 19} and \textbf{20}. When A is vertical and B horizontal, \textbf{lines 22} and \textbf{23}. This commutation is again to win a complex and diverse dataset .
%------------------------------------------------------------------------- 
\Section{Dataset of 2D Maps}
\label{sec_maps}

\begin{figure*}[tb]
    \centering
    \begin{subfigure}[]{0.3\textwidth}
    \includegraphics[width=\linewidth]{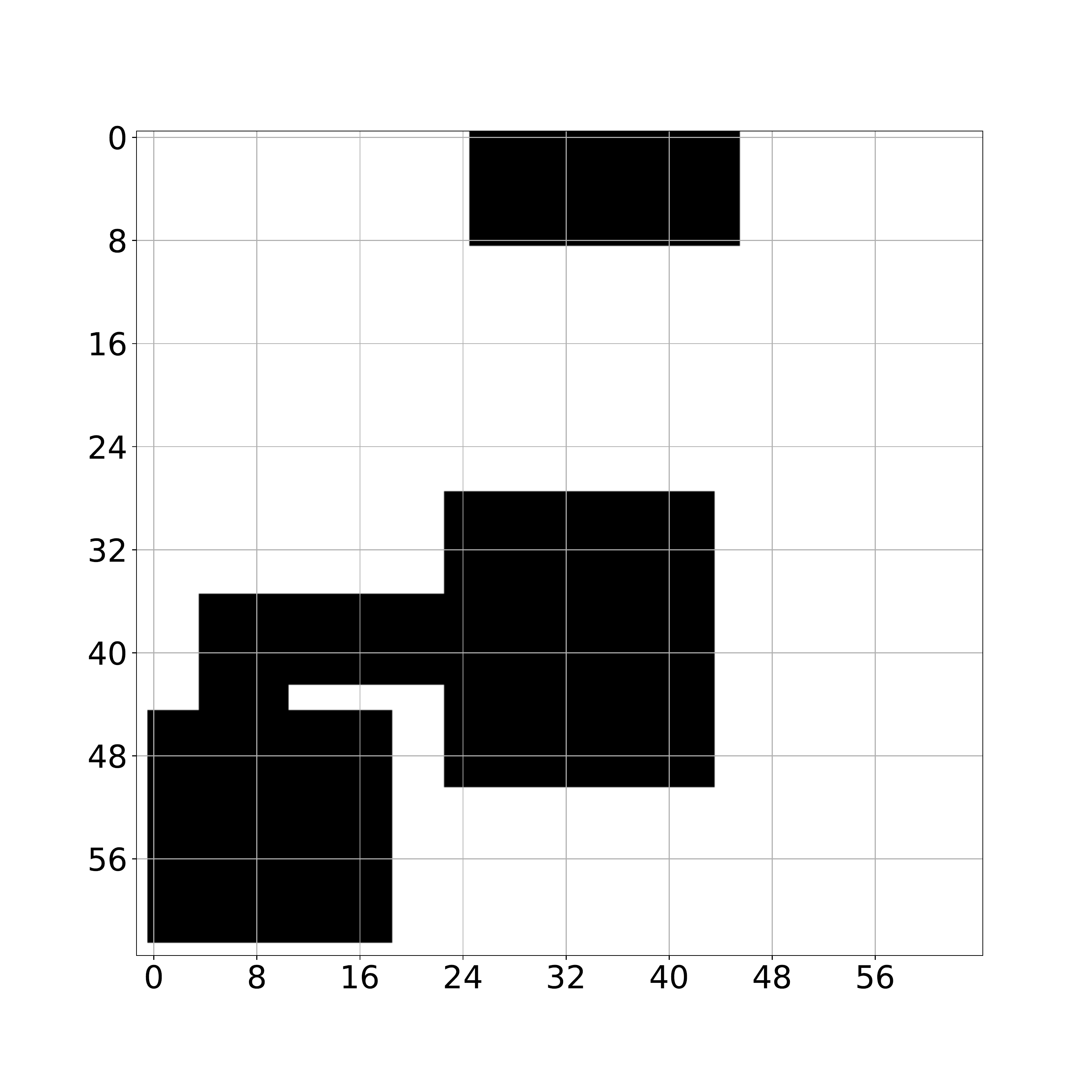}
    \caption{Three rooms with one not connected.}
    \label{map3105}
    \end{subfigure}
    \hspace{0.5cm}
    \begin{subfigure}[]{0.3\textwidth}
    \includegraphics[width=\linewidth]{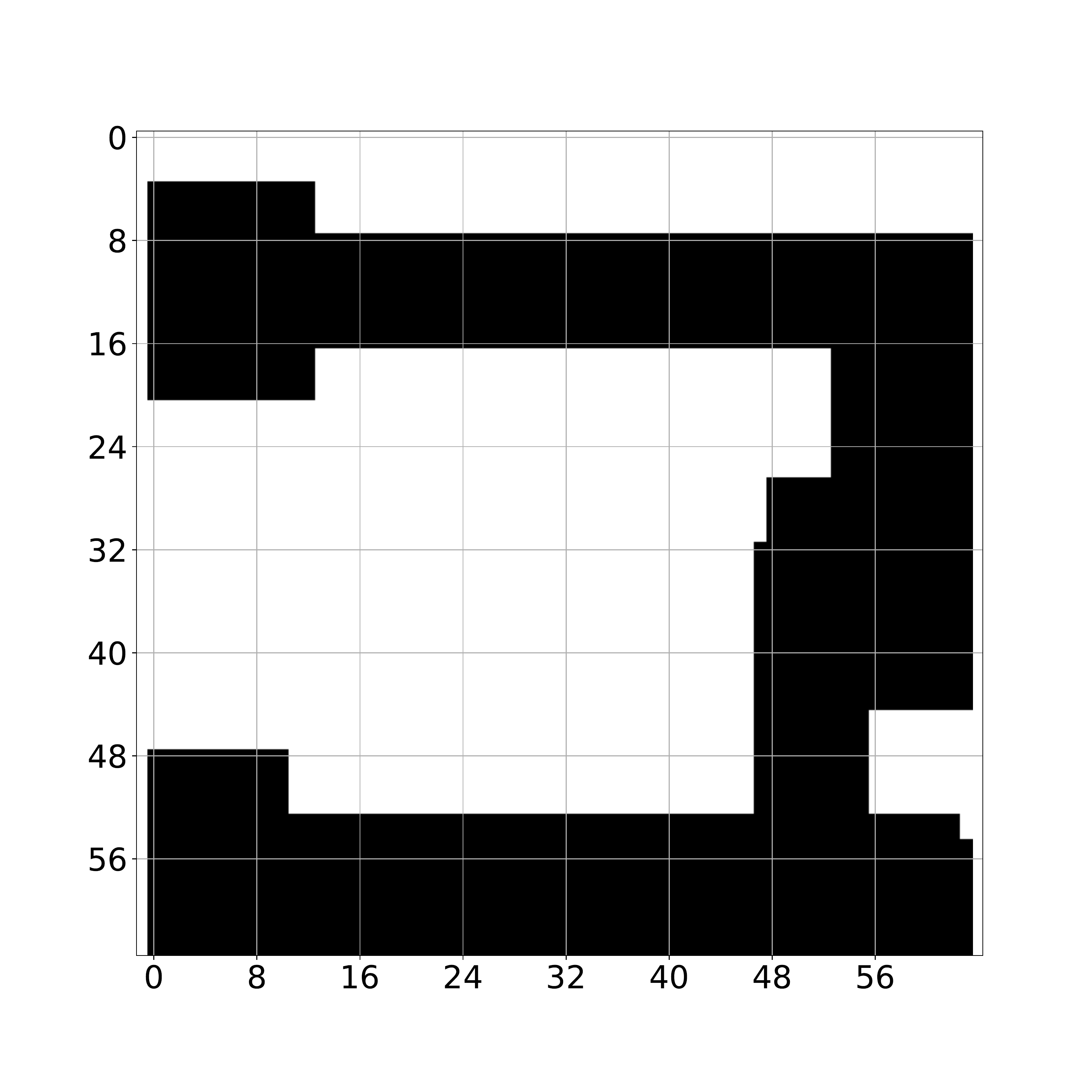}
    \caption{Three rooms all connected.}
    \label{map1762}
    \end{subfigure}
    \hspace{0.5cm}
    \begin{subfigure}[]{0.3\textwidth}
    \includegraphics[width=\linewidth]{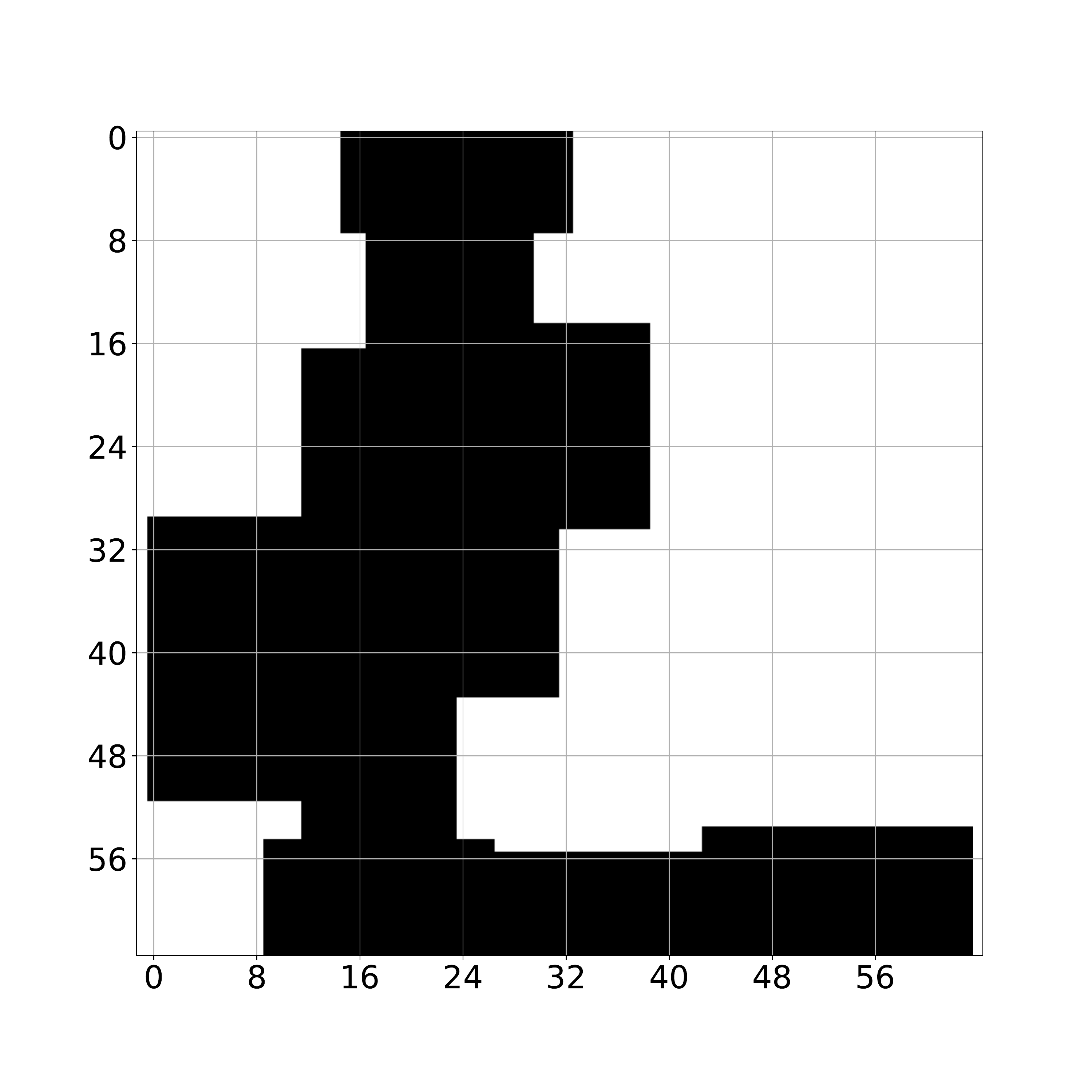}
    \caption{Five rooms.}
    \label{map1401}
    \end{subfigure}
    \begin{subfigure}[]{0.3\textwidth}
    \includegraphics[width=\linewidth]{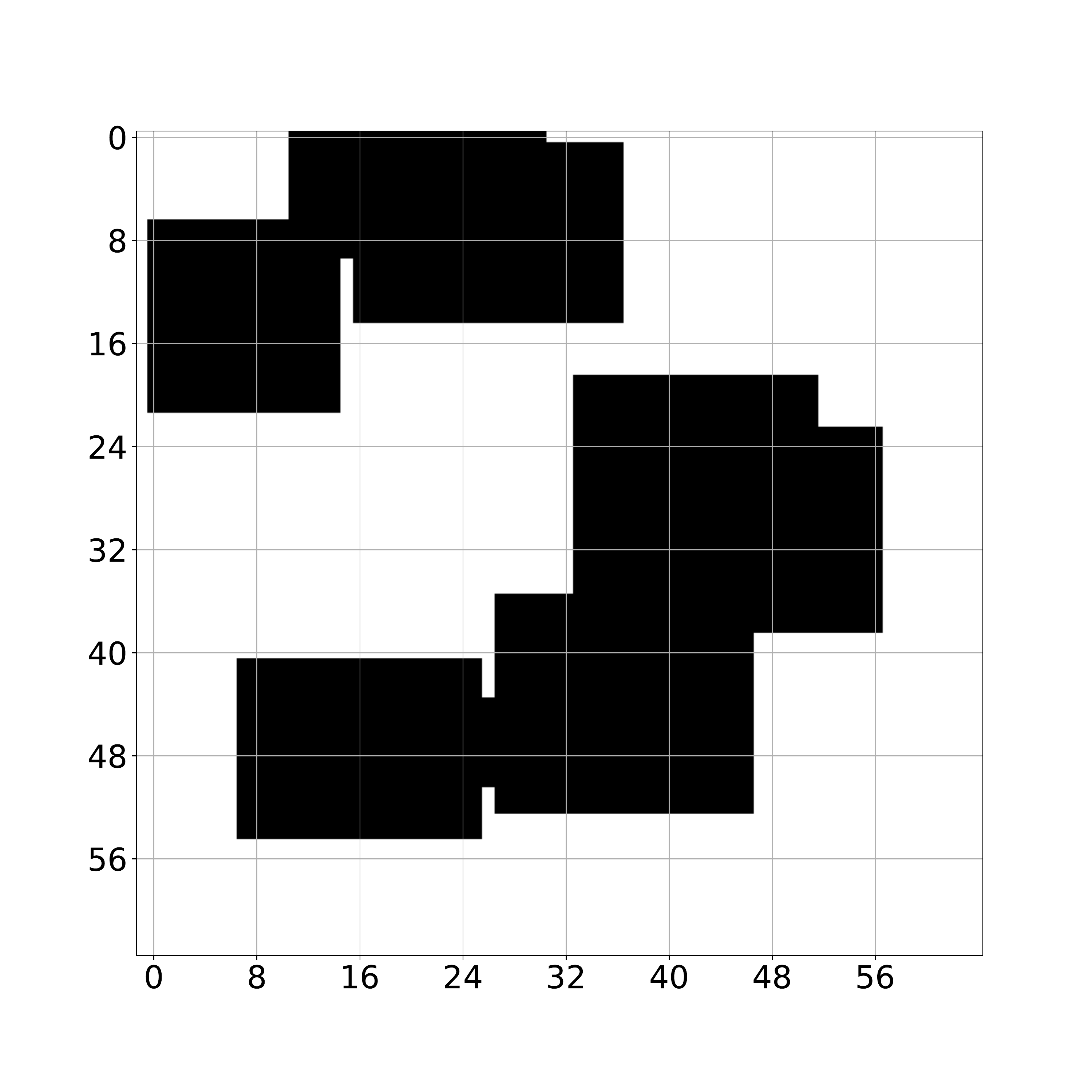}
    \caption{Five rooms with two regions not connected}
    \label{map1842}
    \end{subfigure}
    \hspace{0.5cm}
    \begin{subfigure}[]{0.3\textwidth}
    \includegraphics[width=\linewidth]{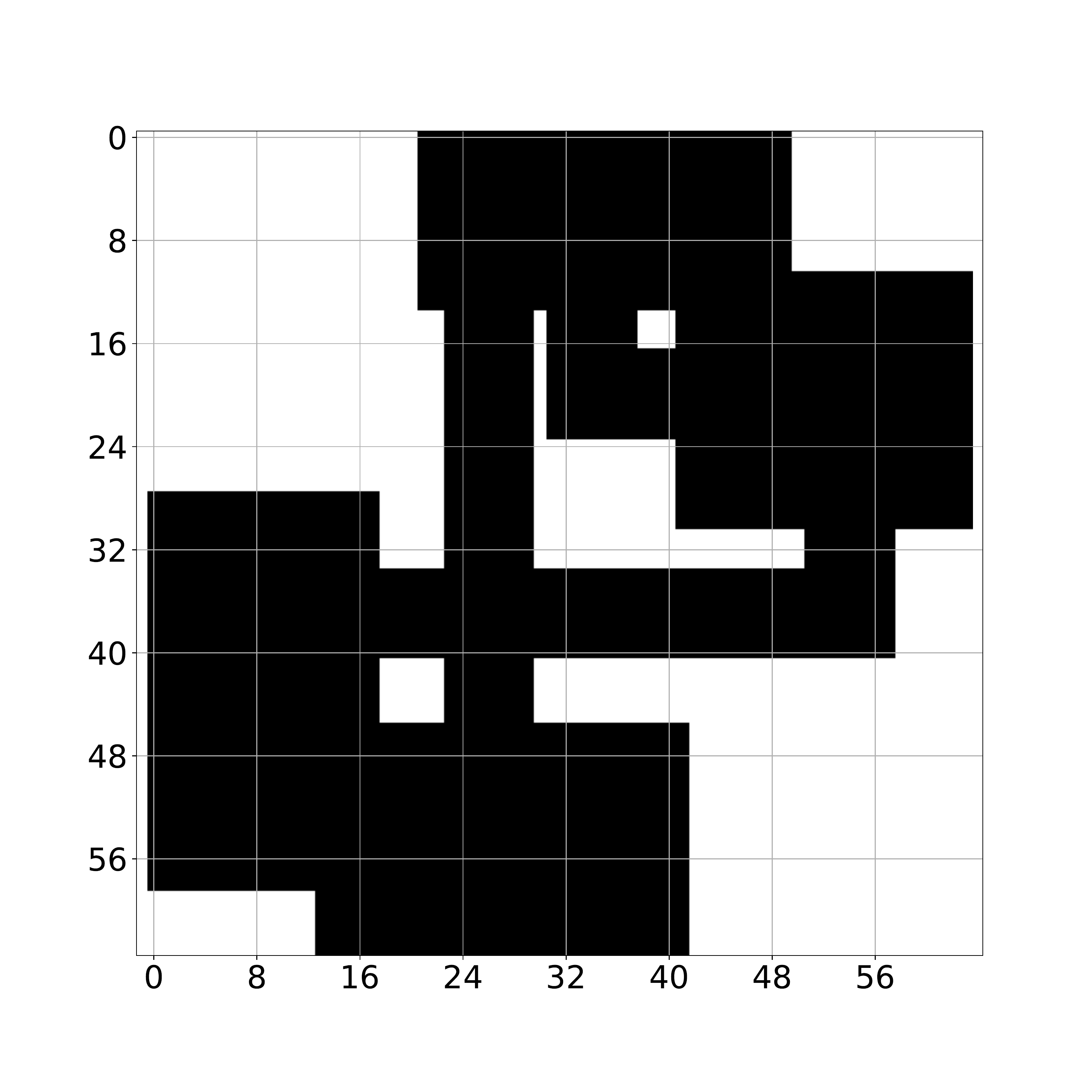}
    \caption{Four rooms with redundant connections.}
    \label{map3297}
    \end{subfigure}
    \hspace{0.5cm}
    \begin{subfigure}[]{0.3\textwidth}
    \includegraphics[width=\linewidth]{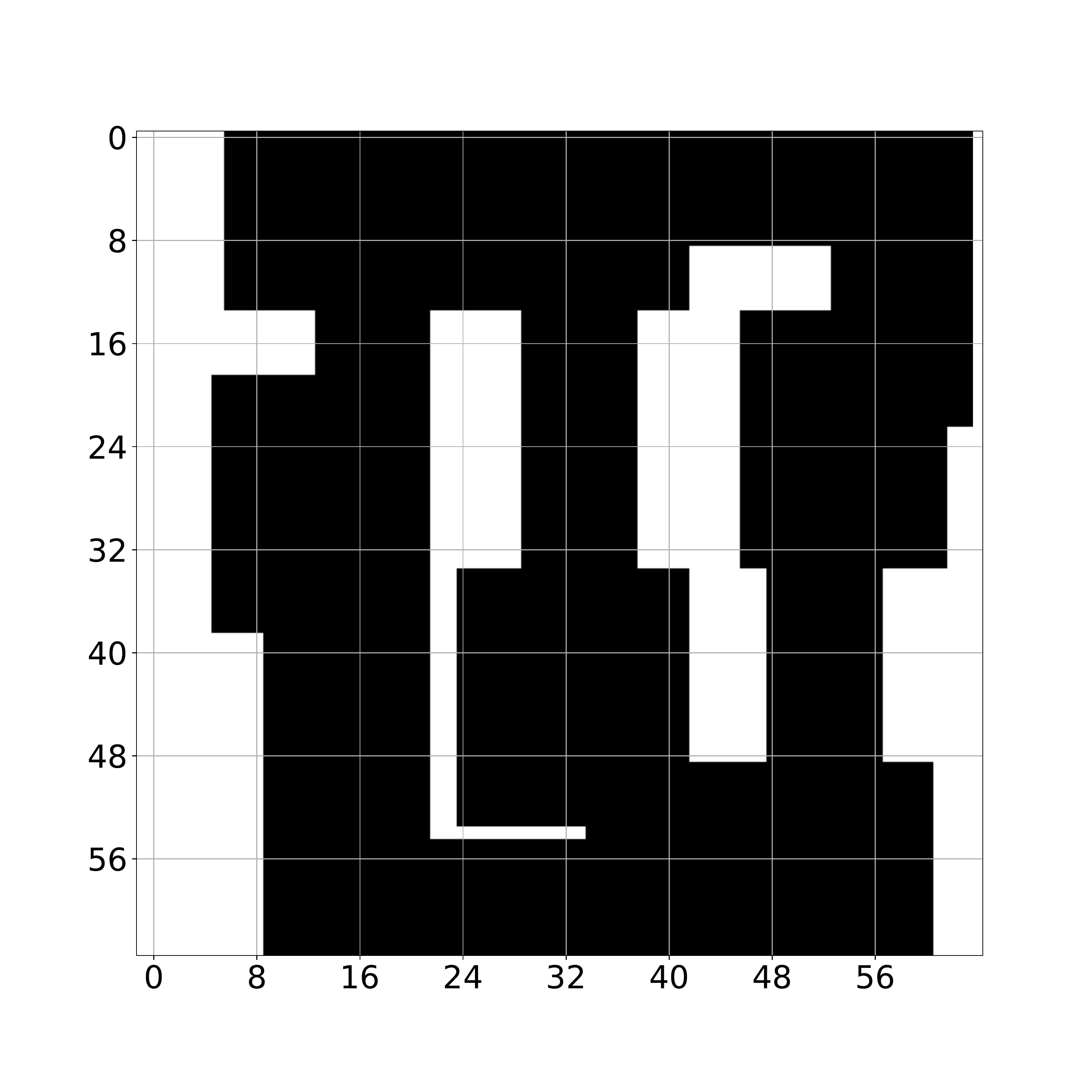}
    \caption{Nine rooms with all the regions connected}
    \label{map961}
    \end{subfigure}
    \caption{Examples of maps included in our dataset. Black color represents the free space and color white represents the occupied space.}
    \label{fig:mapI}
\end{figure*}

In this section, we present a brief description of the generated maps. The dataset  attached, contains 10,000 square maps of 64 pixels for 64 pixels. Each pixel is represented by a value of 1 for occupied space and 0 for free space as shown in Figure \ref{fig:mapI}. The origin is $(0,0)$ and takes place in the upper-left corner and the final pixel position $(63,63)$ is in the opposite corner. 
  
In Figure \ref{fig:mapI}, we show some examples of the maps by the parameters in Table \ref{tab_parameters}. The reader can see in sub-figure \ref{map3105} an example where one room is not connected, this case has the objective to represent the situation where a robot can not access certain spaces. In sub-figure \ref{map961}, we show an example of a map with the maximum number of rooms. 

\begin{table}[tb]
\centering
\begin{tabular}{|l|c|c|}
\hline
\multicolumn{3}{|c|}{\textbf{Parameters for dataset }}                                                                \\ \hline
\textbf{item}    & \multicolumn{2}{c|}{\textbf{value}}                                                   \\ \hline
maps             & \multicolumn{2}{c|}{10,000}                                                           \\ \hline
maps size        & \multicolumn{2}{c|}{64x64 px}                                                         \\ \hline
\textbf{item}    & \multicolumn{1}{l|}{\textbf{lower limit}} & \multicolumn{1}{l|}{\textbf{upper limit}} \\ \hline
rooms            & 2                                         & 9                                         \\ \hline
rooms size row   & 14 px                                     & 33 px                                     \\ \hline
room size column & 14 px                                     & 33 px                                     \\ \hline
passages size    & 7 px                                      & 15 px                                     \\ \hline
\end{tabular}
\caption{Table with parameters for the dataset  associated.}
\label{tab_parameters}
\end{table}

Below we show and describe the parameters for the algorithm. Also, these parameters are saved and provided together with the dataset.
\begin{itemize}
    \item Item: The number of maps. 
    \item Label: A name assigned to identify each map.
    \item Resolution: A relative value to guide the approximate dimension of the map in the real world. 
    \item Passages size: The randomly selected dimension from the lower and upper limits in the Tunnel width limit.
    \item Seed rows: A matrix of one row and $n$ columns with the randomly selected position for rows. 
    \item Seed columns: A matrix of one row and $n$ columns with the randomly selected position for columns. 
    \item Size rows: A matrix of one row and $n$ columns with the randomly selected size length for rooms in pixels.     
    \item Size columns: A matrix of one row and $n$ columns with the randomly selected size width for rooms in pixels.
    \item Direction for passages: A matrix of one row and $2(n-1)$ columns with the sequence of values that indicate the direction of the tunnel.  
    \item Connected: Line matrix with $n-1$ elements. It is a sequence composed of True or False that indicates if room A connects with B (see line 13 \ref{alg:one}).    
\end{itemize}

%------------------------------------------------------------------------
\section{Characterization}
\label{sec_makeup}
In addition to the set of maps represented as binary images, we included in the dataset an associated document characterization. In each map, we calculated the path with two classic algorithms breadth-first search and A* and we provide in a row and column the coordinates of a path that goes through all the seed points of each room, see Figure \ref{fig:map497} for an example, and a connection matrix that show information about the cost or distance in pixels of connecting two rooms.

\begin{itemize}
    \item Row path for breadth-first search: Line matrix with the row coordinate for the path founded.
    \item Column path for breadth-first search: Line matrix with the columns coordinate for the path founded.
    \item Iterations for breadth-first search: The number of iterations required to found a path.  
    \item Connection matrix for breadth-first search: Contain information about the path in pixel between all the points.   
\end{itemize}

\begin{itemize}
    \item Row path for A*: Line matrix with the row coordinate for the path founded.
    \item Column path for A*: Line matrix with the columns coordinate for the path founded.
    \item Iterations for A*: The number of iterations required to found a path.  
    \item Connection matrix for A*: Contain information about the path in pixel between all the points.   
\end{itemize}

%***************************************************** BF and A* 

\begin{figure*}[tb]
    \centering
    \begin{subfigure}[b]{0.45\textwidth}
    \includegraphics[width=\linewidth]{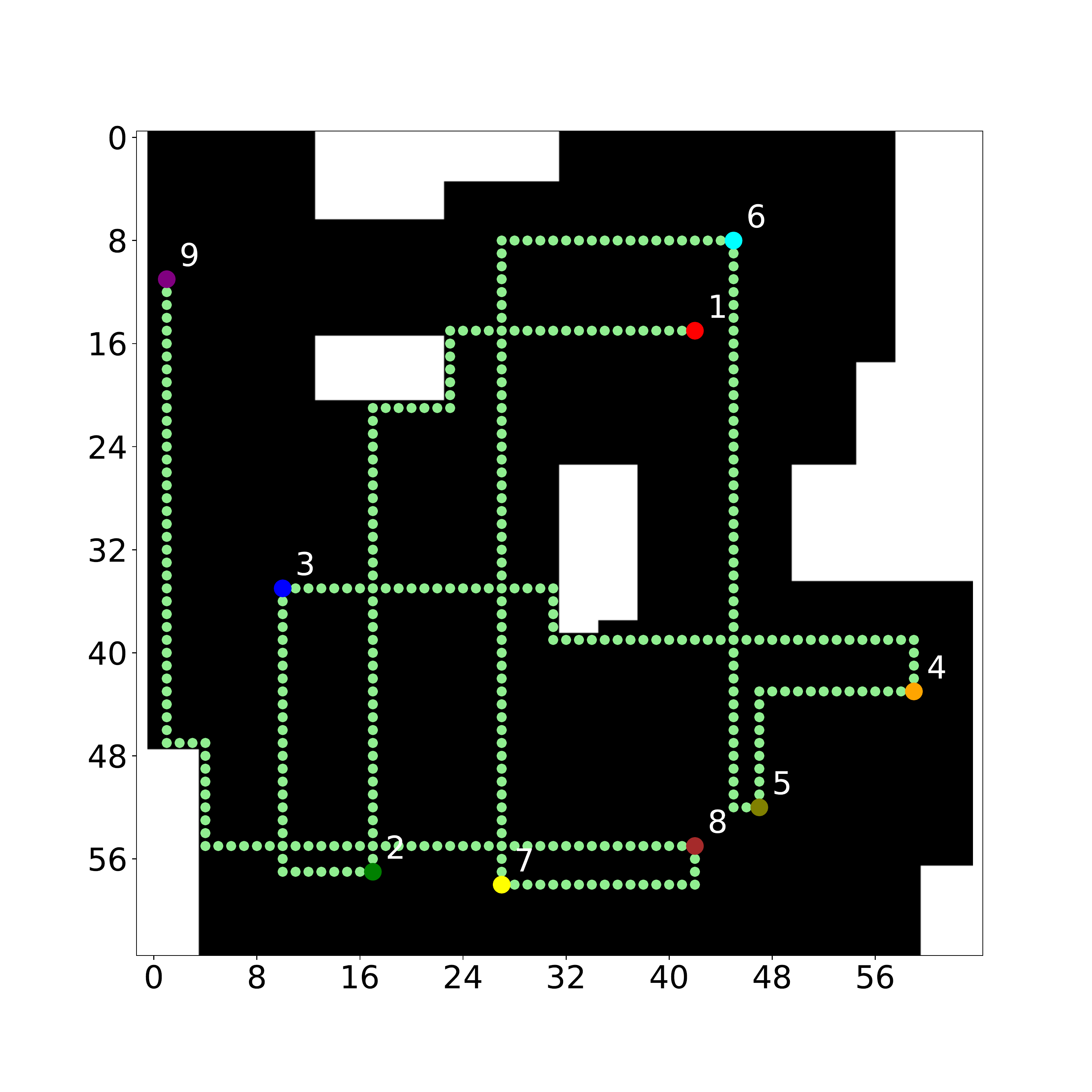}        
    \caption{}
    \label{map_397}
    \end{subfigure}
    \hspace{0.05cm}
    \centering
    \begin{subfigure}[b]{0.45\textwidth}
    \includegraphics[width=\linewidth]{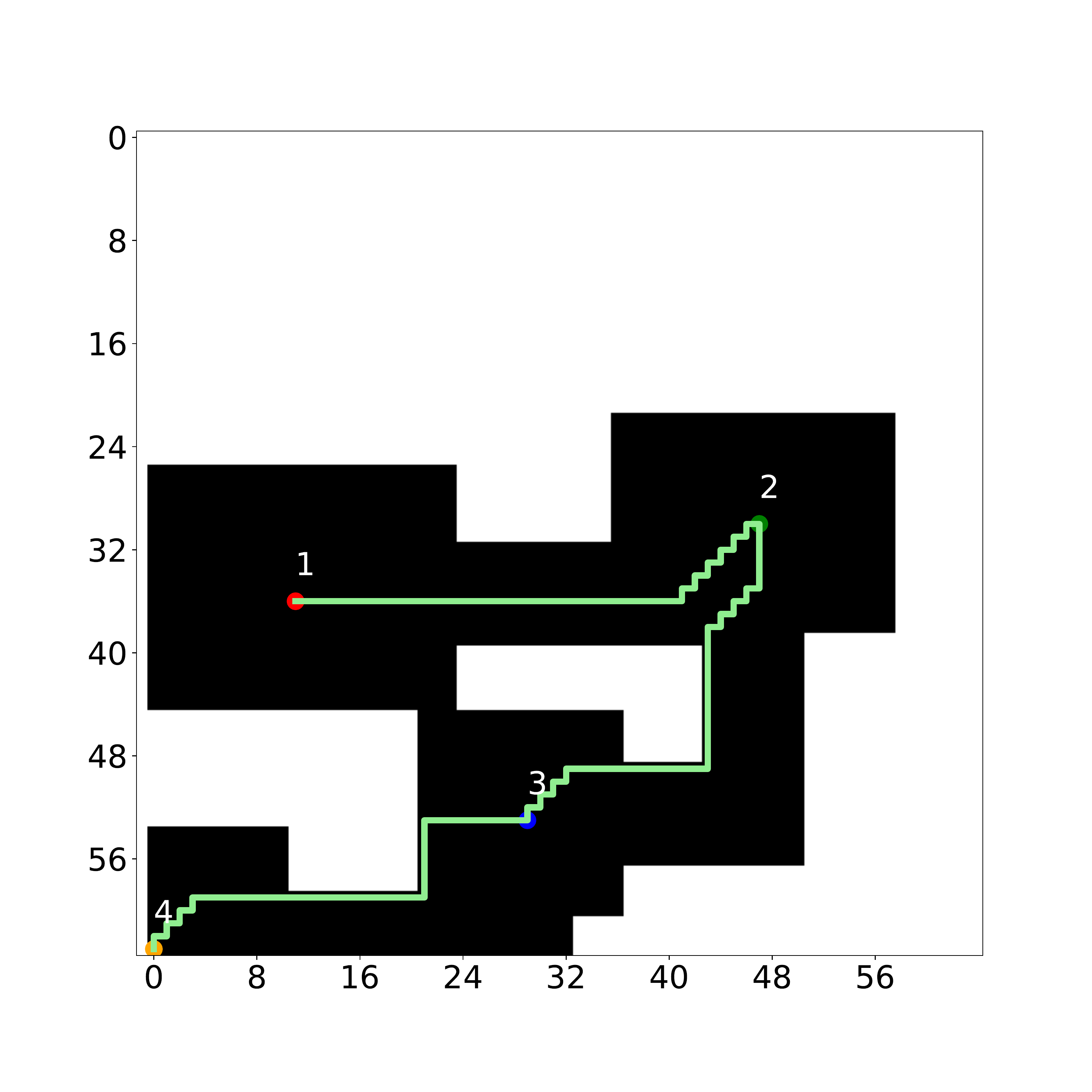}
    \caption{}
    \label{mtx_317}
    \end{subfigure}
    \caption{In sub-figure a, in color light green the path after 1051 iterations for map 417. Path computed with Breath-First search. In sub-figure b, in color light green the path after 16486 iterations for map 397. Path computed with A* search algorithm}
    \label{fig:paths}
\end{figure*}

The path found is the route that connects all the seed points in the original sequences. That means room $1$, room $2$ until room $n$, see Figure \ref{fig:paths}. This is not the optimal path but ensures traffic between all rooms. This path was founded after computing Breadth-first search, one of the simplest and archetype algorithms for finding routes in graphs \cite{IntroductionToAlgorithms}.

\begin{figure}[tb]
    \centering
    \includegraphics[scale=0.70]{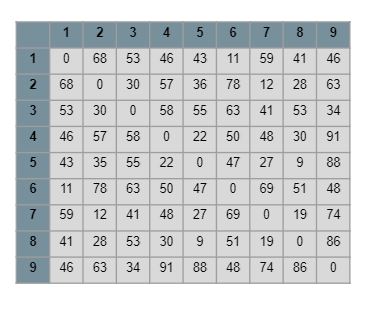}
    \caption{Connection matrix example. In figure the matrix associated to map 417 is drawn.}
    \label{fig:connMtx}
\end{figure}

The connection matrix is a representation where an element $c_{i,j}$ with $i=\{1,2,3.., ...,n\}$ and $j=\{1,2,3.., ...,n\}$ represents the distance in pixels for the $i^{th}$ and $j^{th}$ room. See Figure \ref{fig:connMtx} for some examples. The value of -$1$ is a convention to indicate ``no connection". The value of $0$ appears when it is the same point and the rest values are the path distance or cost in pixels.

%------------------------------------------------------------------------- 

\Section{Conclusions}
\label{sec_conclusion}

We have presented an algorithm for building random 2D grid maps. This algorithm allows building maps of multiple sizes and degrees of complexity by modifying and selecting randomly simple parameters. Along with the algorithm, it is presented a database with 10,000 2D maps using the proposed method. We include together with the dataset, a CSV file with the configuration parameters for each map and their characterization in terms of calculated paths, computed with two classic algorithms. Overall, this information can be useful to train and test machine learning approaches for map-based navigation.

%------------------------------------------------------------------------- 
\Section{Acknowledges}
The authors are grateful for the support provided by CONACYT and the IPN. In addition this work is funded by the SIP-IPN, with registration number 20210268. Jheison Duvier Diaz Ortega is grateful for the support granted in the call for technological development or innovation projects for students of the IPN 2021.
%------------------------------------------------------------
\nocite{ex1,ex2}
\bibliographystyle{latex8}
\bibliography{latex8}

\begin{thebibliography}{10}\setlength{\itemsep}{-1ex}\small

\bibitem{Corke}
P.~Corke.
\newblock {\em Robotics, Vision and Control}.
\newblock Springer, 2017.

\bibitem{PRMRL}
A.~Faust, K.~Oslund, O.~Ramirez, A.~Francis, L.~Tapia, M.~Fiser, and
  J.~Davidson.
\newblock Prm-rl: Long-range robotic navigation tasks by combining
  reinforcement learning and sampling-based planning.
\newblock In {\em 2018 IEEE International Conference on Robotics and Automation
  (ICRA)}, pages 5113--5120, 2018.

\bibitem{Astar}
P.~E. Hart, N.~J. Nilsson, and B.~Raphael.
\newblock A formal basis for the heuristic determination of minimum cost paths.
\newblock {\em IEEE Transactions on Systems Science and Cybernetics},
  4(2):100--107, 1968.

\bibitem{planningLatentSpaces}
B.~Ichter and M.~Pavone.
\newblock Robot motion planning in learned latent spaces.
\newblock {\em IEEE Robotics and Automation Letters}, 4(3):2407--2414, 2019.

\bibitem{PRM}
L.~J. Kavraki L.~E., Svestka~P. and O.~M. H.
\newblock Probabilistic roadmaps for path planning in high-dimensional
  configuration spaces.
\newblock {\em IEEE Transactions on Robotics and Automation}, 12(4):566--580, -
  1996.

\bibitem{lego}
R.~Kumar, A.~Mandalika, S.~Choudhury, and S.~Srinivasa.
\newblock Lego: Leveraging experience in roadmap generation for sampling-based
  planning.
\newblock In {\em 2019 IEEE/RSJ International Conference on Intelligent Robots
  and Systems (IROS)}, pages 1488--1495, 2019.

\bibitem{RRT}
L.~S. M.
\newblock Rapidly-exploring random trees: A new tool for path planning.
\newblock {\em Computer Science Dept, Iowa State University}, -(-):--, - 1998.

\bibitem{dungeons}
S.~N., L.~A., T.~J., L.~R., and B.~R.
\newblock {\em Constructive generation methods for dungeons and levels. In:
  Procedural Content Generation in Games. Computational Synthesis and Creative
  Systems.}
\newblock Springer, 2016.

\bibitem{motionPlanningNetworks}
A.~H. Qureshi, Y.~Miao, A.~Simeonov, and M.~C. Yip.
\newblock Motion planning networks: Bridging the gap between learning-based and
  classical motion planners.
\newblock {\em IEEE Transactions on Robotics}, 37(1):48--66, 2021.

\bibitem{Thrun}
W.~B. Sebastian~Thrun and D.~Fox.
\newblock {\em Probabilistic robotics}.
\newblock MIT Press, 2006.

\bibitem{TowardAutonomous}
F.~C. Shi~Bai and Englot.
\newblock Toward autonomous mapping and exploration for mobile robots through
  deep supervised learning.
\newblock {\em International Conference on Intelligent Robots and Systems},
  -(-):2379--2384, - 2017.

\bibitem{IntroductionToAlgorithms}
R.~L.~R. Thomas H.~Cormen, Charles E.~Leiserson and C.~Stein.
\newblock {\em Introduction to Algorithms}.
\newblock MIT Press, 2001.

\end{thebibliography}
\end{document}